# Second-order Symmetric Non-negative Latent Factor Analysis

Weiling Li, and Xin Luo, *Senior Member, IEEE*

*Abstract*—Precise representation of large-scale undirected network is the basis for understanding relations within a massive entity set. The undirected network representation task can be efficiently addressed by a symmetry non-negative latent factor (SNLF) model, whose objective is clearly non-convex. However, existing SNLF models commonly adopt a first-order optimizer that cannot well handle the non-convex objective, thereby resulting in inaccurate representation results. On the other hand, higher-order learning algorithms are expected to make a breakthrough, but their computation efficiency are greatly limited due to the direct manipulation of the Hessian matrix, which can be huge in undirected network representation tasks. Aiming at addressing this issue, this study proposes to incorporate an efficient second-order method into SNLF, thereby establishing a second-order symmetric non-negative latent factor analysis model for undirected network with two-fold ideas: a) incorporating a mapping strategy into SNLF model to form an unconstrained model, and b) training the unconstrained model with a specially designed second order method to acquire a proper second-order step efficiently. Empirical studies indicate that proposed model outperforms state-of-the-art models in representation accuracy with affordable computational burden.

## I. INTRODUCTION

WITH the explosive emergence of big data scenarios, undirected network become a common and important data mode to describe the quantifiable interrelationship within a large set of entities, like associations between users in social networks [1-3] and relations between proteins in molecular biology [4, 5]. An undirected network can be denoted by a symmetric matrix. As a result of massive associated entities and cost constrained observations, these matrices are usually Symmetric, High-Dimensional and Incomplete (SHDI). For better understanding the relation pattern with in an entity set by analyzing the very limited number of observations, a well-designed representation learning model is highly desired.

A nonnegative matrix factorization (NMF) model has proven to be efficient in extracting useful information from relation matrix by factorizing each row and column of target matrix into low dimensional non-negative latent factor (NLF) spaces [6-8]. Multiple NMF models are designed for symmetric and non-negative input, e.g., symmetric and non-negative matrix factorization (SNMF) model [9] and symmetric constrained NMF (SymNMF) model [10]. However, the NMF models are design for representing full matrix, recall that in large-scale undirected network representation problems, the target incomplete matrices are extremely sparse, analyzing models require well-designed strategies to process the input matrices efficiently.

To analyze an incomplete matrix, a possible way is to fill its unknown elements with assumptions firstly. The Expectation-Maximization (EM) model assumes that the unknown entries obey a normal distribution hence to estimate their values based on this assumption. The Weighted Non-negative Matrix Factorization (WNMF) employs a binary parameter, which is set to one if the current entry is not empty and zero otherwise [6]. However, unreliable estimations will lead to inaccurate representation results. To avoid uncertain estimation for SHDI input, Luo et al. propose the symmetric non-negative latent factor (SNLF) model bases on the idea of single element dependence [11], which is able to implement efficient extraction of NLFs for analyzing incomplete data. With the insight of SNLF model, the symmetric non-negative latent factor models are proposed [12, 13]. These models are high efficiency for analyzing undirected network and inspire some models [14-17].

Recall that obtaining optimal NLFs through a SNLF model is a non-convex optimization problem. For fulfil such a task, the line search framework is a common solution. Determining a search direction with current point information is the core for implementing line-search-based methods. The information can be first-ordered or second-ordered. Note that the curve of a high-dimensional and non-convex function is very complex, the performance of first-order-based methods, which generate their searching direction with gradient information and focus on the first order stationary point, are easily affected by saddle points [18, 19]. Hence, for a SNLF model, second-order-based methods are more suitable for obtaining optimal NLFs [20, 21]. It gives an insight that the representation ability of SNLF models will improve if we can utilize the second order information properly to guide the training process. Unfortunately, most SNLF models are first-order based [12-15]. The reason for the absence of the second-order-based SNLF model is threefold:

a) The classical second-order-based method, i.e., Newton-type method, acquires the update, e.g., $\Delta x$, by solving following equation directly

$$H_f(x)\Delta x + \nabla f(x) = 0, \qquad (1)$$

where $\nabla f(x)$ and $H_f(x)$ denote function $f(x)$'s gradient and Hessian matrix, respectively. It requires $\Theta(|x|^3)$ to compute the inverse of the Hessian matrix and $\Theta(|x|^2)$ to store the result [22, 23]. It is unaffordable in practical undirected network analysis jobs since $|x|$ can be extremely huge. Moreover, improved Newton methods like quasi-Newton methods also suffer from high computational cost in large-scale network representation problems;

b) The non-negative constraints will disturb the process of Hessian generation since the output latent factors play the role of decision parameters simultaneously;

c) The objective function of an SNLF model is high-dimensional and non-convex, which makes it difficult to determine the positivity and singularity of its Hessian. Therefore, the Newton iteration $\Delta x$ may not be a descent direction [24] and has negative effects on model's convergence. Moreover, this problem also limits the use of some improved Newton methods.

To overcome aforementioned issues, this paper proposes a <u>S</u>econd-order <u>S</u>ymmetric <u>N</u>on-negative <u>L</u>atent <u>F</u>actor Analysis (S²NLF) model. It adopts implicit symmetry and non-negative constraint modeling method by separating the output NLFs from the decision parameters with a mapping function that can always maintain the non-negativity of NLFs. Its parameters are naturally constrained without additional constraints. Hence its NLFs can be obtained by utilizing unconstrained second order methods. Instead of solving (1) exactly, an adjustable vector, e.g., $\kappa$, makes it possible to solve (2) with more flexible methods,

$$H_f(x)\Delta x + \nabla f(x) = \kappa. \quad (2)$$

Conjugate Gradient (CG) [25, 26, 29, 30] is an effective method for solving (2) with lower computational cost, e.g., $\Theta(|x|^2)$. Moreover, by replacing the Hessian matrix with a Gauss-Newton approximation and incorporating a Hessian-vector strategy, proposed model avoid the negative curvature problem and reduces the complexity of calculation, which makes it more practical in analyzing high-dimensional and incomplete data. The main contributions of this work include:
a) incorporating biases and a mapping strategy into SNLF model to form an unconstrained model with higher representation ability;
b) A novel and effective second order-based method is designed for proposed unconstrained SNLF model for acquiring more accurate representation results;
c) Detailed theoretical convergence analysis of proposed algorithm and its complexity;

Empirical studies on five real SHDI undirected networks indicate that proposed model performs well in representing large-scale undirected networks with affordable computational burden. The remainder of this paper is organized as follows: Section II gives the preliminaries; Section III presents the details of the proposed model; Section IV states the empirical studies; and finally, Section V concludes this paper.

## II. PRELIMINARIES

Given an undirected network composed of $V$ entities, which can be described by a matrix $G^{|V|\times|V|}$, an SNLF model takes G as its fundamental input [12, 13] and seeks for NLF matrix $A^{|V|\times d}$ with $d\ll|V|$ based on G's known entry set $E$ [31, 48-50] to build G's rank-$d$ approximation:

$$\hat{G} = AA^T \Rightarrow \hat{g}_{u,i} = \sum_{m=1}^{d} a_{u,m} a_{i,m} \quad (3)$$

where $\hat{g}_{u,i}$ denotes the approximation to $g_{u,i}\in E$, $a_{u,n}$ and $a_{i,n}$ are specified NLFs in $A$.

To obtain optimal $A$, a learning objective is needed. This study adopts the Euclidean distance between G and $\hat{G}$ on $E$ for such a purpose:

$$\arg\min_{A} S(A) = \frac{1}{2}\sum_{g_{u,i}\in E}\left(g_{u,i} - \sum_{m=1}^{d} a_{u,m} a_{i,m}\right)^2 \quad (4)$$

$$s.t. \ \forall u,i \in V, m\in\{1,\ldots,d\}: a_{u,n}\geq 0, a_{i,n}\geq 0.$$

yet theories and methods achieved in this study are highly compatible with other learning objectives. Note that the learning objective (4) is bilinear and non-convex.

## III. AN S²NLF MODEL

### A. Learning Objective

It is clear that the additional non-negative constraints will disturb the second order optimization process and lead to inaccurate representation results. To address this issue, separating the output NLFs from the decision parameters is a possible solution. Hence, we introduce $x$, a decision parameter vector of size $|V|\times d$, where each NLF in $A$ is the output of an element in $x$ by a sigmoid function [32-34], e.g., $\phi(x)=(1+e^{-x})^{-1}$:

$$a_{(u)m} = \phi(x_{(u)m}), a_{(i)m} = \phi(x_{(i)m}). \quad (5)$$

Then, constrains in objective function (4) can be removed:

$$\arg\min_{x} S(x) = \frac{1}{2}\sum_{g_{u,i}\in E}\left(g_{u,i} - \sum_{n=1}^{d} \phi(x_{(u)n})\phi(x_{(i)n})\right)^2 \quad (6)$$

As discussed in prior works, integrating biases into a latent factor analysis-based model can further improve its performance [18, 39]. It inspires us to incorporate biases into SNLF's objective for improve its representation ability. An SNLF incorporates biases as follows:

$$\arg\min_{A,B} S(A,B) = \frac{1}{2}\sum_{g_{u,i}\in E}\left(g_{u,i} - b_u - b_i - \sum_{n=2}^{d} a_{(u)n}a_{(i)n}\right)^2,$$

$$s.t. \ \forall u,i \in V, m\in\{1,\ldots,d\}: \quad (7)$$

$$b_u \geq 0, b_i \geq 0, a_{u,n}\geq 0, a_{i,n}\geq 0.$$

where $B$ is a length-$|V|$ vector, $b_u, b_i\in B$. To remove the non-negative constraints for $B$ from (20), we make each NLF in $B$ to be the output of an element in $x$ by a sigmoid function:

$$b_u = \phi(x_{(u)1}), b_i = \phi(x_{(i)1}). \quad (8)$$

Hence, we have the S²NLF model as follows:

$$\arg\min_{x} S(x) = \frac{1}{2}\sum_{g_{u,i}\in E}\Big(g_{u,i} - \phi(x_{(u)1}) - \phi(x_{(i)1})$$

$$- \sum_{n=2}^{d}\phi(x_{(u)n})\phi(x_{(i)n})\Big)^2. \quad (9)$$

### B. Second-order Learning Scheme

To be a specific line search optimizer, CG approximates to the minimization by the following update rule:

$$x_{k+1} = x_k + \alpha_k p_k \quad (10)$$

where $\{p_1, p_2, \ldots, p_k, \ldots\}$ is said to be conjugate with respect to the symmetric positive definite matrix, e.g., $H_S(x)$, and $\alpha_k$ is given explicitly by

$$\alpha_k = -\frac{\kappa^T p_k}{p_k^T H(x_k) p_k} \quad (11)$$

By obtaining an approximation of Newton step, e.g., $\Delta x$, with Conjugate Gradient (CG) [25, 26], an SNLF model can acquire desired $A$ finally. Note that CG is an iterative solver capable of solving the linear system (2) approximately without inversing the Hessian matrix $H_S(x)$. However, it requires $H_S(x)$ and the computational complexity, i.e., $\Theta((|x|)^2)$ is still unaffordable for large-scale undirected network. Hence, this work incorporates a Hessian-vector strategy into CG to reduce is computational cost. From (11), it is clear that in CG, $H_S(x)$ appears with a vector, e.g., $p$ and the calculation rely on the whole product $\omega_S=H_S(x)p$ but not $H_S(x)$ alone. Since $\omega_S$ actually denotes $S$'s directional derivate along vector $p$'s direction,:

$$\omega_S = H_S(x)p = \frac{\partial(\nabla S(x+\xi\cdot p))}{\partial\xi}\bigg|_{\xi\to 0}. \quad (12)$$

Specifically, $\omega_S$ can be derived by taking the R-operation [27, 28] on the gradient $\nabla S(x)$. Note that R-operation

follows the usual rules of differentiation:

$$a.\ \mathrm{R}(\boldsymbol{x}) = \boldsymbol{p};$$
$$b.\ \mathrm{R}(f(\boldsymbol{x})) = \mathrm{R}(\boldsymbol{x})f'(\boldsymbol{x}); \quad (13)$$
$$c.\ \mathrm{R}(f(\boldsymbol{x}) + g(\boldsymbol{x})) = \mathrm{R}(f(\boldsymbol{x})) + \mathrm{R}(g(\boldsymbol{x})).$$

Thus, $\boldsymbol{\omega}_S$ is computed as follows:

$$\boldsymbol{\omega}_S = \left( \mathrm{R}\left( \frac{\nabla S(\boldsymbol{x})}{x_1} \right), \ldots, \mathrm{R}\left( \frac{\nabla S(\boldsymbol{x})}{x_{|V| \times d}} \right) \right). \quad (14)$$

Then, (2) can be solved at an affordable cost. Recall that the CG method is designed to solve positive definite systems, and the matrix $H_S(\boldsymbol{x})$ may have negative eigenvalues. According to [24], the negative curvature problem can be avoided by replacing $H_S(\boldsymbol{x})$ with its Gauss-Newton approximation, i.e., $N_S(\boldsymbol{x})$:

$$N_S(\boldsymbol{x}) \Delta \boldsymbol{x} + \nabla S(\boldsymbol{x}) = \kappa. \quad (15)$$

To obtain $N_S(\boldsymbol{x})$, we firstly regard the objective function $S(\boldsymbol{x})$ as a composition of two functions, e.g., $S(\boldsymbol{x}) = B(F(\boldsymbol{x}))$, and ensure that $B()$ is convex. Let $W = F(\boldsymbol{x})$, we have

$$w_{u,i} = \phi(x_{(u)1}) + \phi(x_{(i)1}) + \sum_{m=2}^{d} \phi(x_{(u)m}) \phi(x_{(i)m}), \quad (16)$$

where $w_{u,i}$ denotes the element of $W$. Incorporating (16) into an S²NLF model, we have:

$$S(W) = \frac{1}{2} \sum_{g_{u,i} \in E} (g_{u,i} - w_{u,i})^2. \quad (17)$$

Then, the product of an arbitrary $v$ and $N_S(\boldsymbol{x})$, e.g., $\omega_N$, can be derived as follows,

$$\omega_N = N_S(\boldsymbol{x})\boldsymbol{p} = J_F(\boldsymbol{x})^{\mathrm{T}} J_F(\boldsymbol{x})\boldsymbol{p}, \quad (18)$$

by noting that $S(W)$'s second-order derivative is the identity matrix [27, 28] and $J_F(\boldsymbol{x})$ is the Jacobian matrix of $F(\boldsymbol{x})$, the following equation holds,

$$J_F(\boldsymbol{x})\boldsymbol{p} = \left. \frac{\partial(\nabla F(\boldsymbol{x} + \xi \boldsymbol{p}))}{\partial \xi} \right|_{\xi \to 0} = R(F_{(u,i)}(\boldsymbol{x}))\Big|_{(u,i) \in E}. \quad (19)$$

According to the nature of Jacobian matrices, we have:

$$J_F(\boldsymbol{x}) = \left( \left. \frac{\partial F_{(u,i)}(\boldsymbol{x})}{\partial \boldsymbol{x}} \right|_{(u,i) \in E} \right). \quad (20)$$

By combining (16) and (19), we have

$$J_F(\boldsymbol{x})\boldsymbol{p} = \phi'(x_{(u)1})\boldsymbol{p}_{(u)1} + \phi'(x_{(i)1})\boldsymbol{p}_{(i)1} + \sum_{m=2}^{d} \big( \phi'(x_{(u)m}) \\ \times \boldsymbol{p}_{(u)m}\phi(x_{(i)m}) + \phi(x_{(u)m})\phi'(x_{(i)m})\boldsymbol{p}_{(i)m} \big). \quad (21)$$

By combining (21) and (20), $\omega_G$ can be derived as follows:
$\forall u \in V$, $n = 1$:

$$\omega_{N(u)n} = \sum_{i \in E(u)} \Big( \phi'(x_{(u)1}) \times \big( \phi'(x_{(u)1})\boldsymbol{p}_{(u)1} \\ + \phi'(x_{(i)1})\boldsymbol{p}_{(i)1} + \sum_{m=2}^{d} \big( \phi'(x_{(u)m}) \times \boldsymbol{p}_{(u)m}\phi(x_{(i)m}) \\ + \phi(x_{(u)m})\phi'(x_{(i)m})\boldsymbol{p}_{(i)m} \big) \big) \Big),$$

$\forall u \in V$, $n = 2 \sim d$:

$$\omega_{N(u)n} = \sum_{i \in E(u)} \Big( \phi'(x_{(u)n})\phi(x_{(i)n}) \times \big( \phi'(x_{(u)1})\boldsymbol{p}_{(u)1} \\ + \phi'(x_{(i)1})\boldsymbol{p}_{(i)1} + \sum_{m=2}^{d} \big( \phi'(x_{(u)m}) \times \boldsymbol{p}_{(u)m}\phi(x_{(i)m}) \\ + \phi(x_{(u)m})\phi'(x_{(i)m})\boldsymbol{p}_{(i)m} \big) \big) \Big). \quad (22)$$

## C. Integrating Damping and Regularization

Note that the theoretical requirement of proximity is a foundation of quadratic convergence of second-order-based methods. However, for a large-scale non-convex optimization problem, e.g., building SNLF models, random initialization may not close enough to a local minimum. Hence, an effective damping method is important to improve its performance [35]. Utilizing Tikhonov damping is an appropriate solution when the quadratic model is most untrustworthy along directions of very low-curvature. In this study, we utilize this technique by replacing $N_S(\boldsymbol{x})$ with $D(\boldsymbol{x})$ as follows:

$$\begin{cases} D(\boldsymbol{x}) \Delta \boldsymbol{x} + \nabla S(\boldsymbol{x}) = \boldsymbol{\kappa}, \\ D(\boldsymbol{x}) = N_S(\boldsymbol{x}) + \mu \cdot \mathbf{I}. \end{cases} \quad (23)$$

where $\mu$ denotes a positive constant and $\mathbf{I}$ an identity matrix. Then, the ultimate matrix-vector product turns to:

$$\boldsymbol{\omega}_D = D(\boldsymbol{x}) \cdot \boldsymbol{p} = (N_S(\boldsymbol{x}) + \mu \cdot \mathbf{I})\boldsymbol{p} = \boldsymbol{\omega}_N + \mu \cdot \boldsymbol{p}. \quad (24)$$

Moreover, an SHDI matrix is extremely sparse, it makes overfitting a commonly encountered situation in building SNLF models and it largely reduces the models' performance. To avoid overfitting, regularization is a promising method [36, 37]. Therefore, we associate objective (9) with the Tikhonov regularizing term [38]:

$$\arg\min_{\boldsymbol{x}} Z(\boldsymbol{x})$$
$$= \frac{1}{2} \sum_{g_{u,i} \in E} \left( \Big( g_{u,i} - \phi(x_{(u)1}) - \phi(x_{(i)1}) - \sum_{n=2}^{d} \phi(x_{(u)n}) \right. \quad (25)$$
$$\left. \phi(x_{(i)n}) \Big)^2 + \lambda \left( \sum_{n=1}^{d} \phi^2(x_{(u)n}) + \sum_{n=1}^{d} \phi^2(x_{(i)n}) \right) \right).$$

and the effect of regularizing term is controlled by the parameter $\lambda$. Based on (14), the Hessian-vector product of $Z(\boldsymbol{x})$ can be derived as follows:

$$H_Z(\boldsymbol{x})\boldsymbol{p} = \left( R\left( \frac{\partial Z(\boldsymbol{x})}{x_1} \right), \ldots, R\left( \frac{\partial Z(\boldsymbol{x})}{x_{|V| \times d}} \right) \right). \quad (26)$$

According to (25) and (9), it is clear that the matrix-vector of Tikhonov regularizing term, e.g., $\Delta$, depends on the difference between $Z(\boldsymbol{x})$ and $S(\boldsymbol{x})$:

$$H_Z(\boldsymbol{x})\boldsymbol{p} - H_S(\boldsymbol{x})\boldsymbol{p} \Rightarrow$$
$$\forall u \in V,\ m = 1 \sim d: \quad (27)$$
$$\Delta_{(u)m} = \lambda \phi'(x_{(u)m})\phi(x_{(u)m})\boldsymbol{p}_{(u)m}|E(u)|.$$

where $|E(u)|$ denotes the number of known edges on $u$th row. Hence, $\omega_E = G_E(\boldsymbol{x})\boldsymbol{p}$ can be derived as follows:

$$\boldsymbol{\omega}_Z = N_Z(\boldsymbol{x})\boldsymbol{p} \approx N_S(\boldsymbol{x})\boldsymbol{p} + (H_Z(\boldsymbol{x})\boldsymbol{p} - H_S(\boldsymbol{x})\boldsymbol{p});$$
$$\forall u \in V,\ m = 1 \sim d: \quad (28)$$
$$\boldsymbol{\omega}_{Z(u)m} = \boldsymbol{\omega}_{N(u)m} + \lambda \phi'(x_{(u)m})\phi(x_{(u)m})\boldsymbol{p}_{(u)m}|E(u)|.$$

## IV. EXPERIMENTS

In order to evaluate the representation performance of SNLF models for large-scale undirected networks, to predict the missing data of an SHDI matrix is a common method [39, 41]. By measuring the gap between recovered data and known data, the effectiveness of low rank representation features can be measured. Hence, we adopt the task of missing data prediction as the first evaluation protocol to validate involved models' performance.

## A. Experiment Settings

**Evaluation Metrics.** In order to quantitatively measure the effect of missing data completion, this study adopts Root mean squared error (RMSE) [42-44]:

$$RMSE = \sqrt{\left(\sum_{g_{u,i} \in E}(g_{u,i} - \hat{g}_{u,i})^2\right) / |L|}, \quad (29)$$

where $L$ denotes the testing dataset disjoint with $E$.

**Data sets.** The experiments are conducted on five SHDI datasets, whose details are given below:
a) D1: Homo Sapiens dataset from STRING [40]. It consists of 1,182,124 interactivities among 5,194 proteins, its data density is 4.38%.
b) D2: Emericella nidulans dataset from STRING [40]. It consists of 1,120,028 interactivities among 7,963 proteins, its data density is 1.77%.
c) D3: Cellulophaga lytica dataset from STRING [40]. It consists of 638,904 interactivities among 3,277 proteins, its data density is 5.95%.
d) D4: Netscience dataset. It is a network describing a co-authorship of scientists. It consists of 5,484 relations among 1,589 scientists. Its data density is 0.22%.
e) D5: Invisible knowledge network. It consists of 57,022 relations among 2,427 subjects. Its data density is 0.97%.

In order to evaluate $S^2NLF$'s performance on different data density, for each data set, we generate two data cases following the data partition rule given in Table II.

TABLE II
ADOPTED DATA CASES IN THE EXPERIMENTS

| Data Set No. | Train:Test | |E| | |L| |
|---|---|---|---|
| C1.1 | 20:80 | 236425 | 945699 |
| C1.2 | 50:50 | 591062 | 591062 |
| C2.1 | 20:80 | 224006 | 896022 |
| C2.2 | 50:50 | 560014 | 560014 |
| C3.1 | 20:80 | 127781 | 511123 |
| C3.2 | 50:50 | 319452 | 319452 |
| C4.1 | 20:80 | 1097 | 4387 |
| C4.2 | 50:50 | 2742 | 2742 |
| C5.1 | 20:80 | 11404 | 45618 |
| C5.2 | 50:50 | 28511 | 28511 |

The detailed evaluation process is as follows,
a) Each SHDI matrix is divided into 10 pieces equally, and restore $E$ and $L$ following the rule given in table II. We repeat the cross-validation process on each data case for 10 times, where each time all models are initialized with the same randomly-generated arrays, thereby reducing the impact by initial hypothesis.
b) On each fold, we take 90% of $E$ to train the model and the remaining 10% of $E$ as the validation set to monitor the training process.
c) In detail, the training process terminates if the prediction error tested on the validation set rises, or the error difference between two consecutive iterations becomes smaller than $10^{-5}$ for 10 continuous iterations, or the consumed iteration count exceeds a preset threshold, i.e., 500.

**Implementation Details.** All experiments are executed on a bare machine with Intel-i5 2.5 GHz CPU and 32GB RAM. JAVA-SE-7U60 is used to implement tested models.

## B. Comparison against State-of-the-Art Approaches

**Compared Models.** Six models are involved in this set of experiments, whose details are listed below.

**M1. $S^2NLF$.** A second-order symmetric non-negative latent factor model proposed in this work.

**M2. AMF** [45]. An alternating direction method-based matrix factorization model. It represents the target matrix with matrix factorization-based method and adopts the principle of the alternating direction method (ADM) to accelerate the model convergence.

**M3. SNLF** [12]. A symmetric and non-negative latent factor (SNLF) model which is equipped high efficiency, non-negativity, and symmetry.

**M4. BSNLF** [13]. A biased symmetric and non-negative latent factor model which integrates linear biases into the SNLF model for higher performance.

**M5. SymANLS** [46]. A symmetric non-negative matrix factorization model that introduces a regularization to make its learning objective symmetry-aware, and then minimizes the learning objective via an alternating non-negative least squares algorithm.

**M6. SIF** [47]. A neural network-based latent factor analysis model. It adopts a neural interaction function and multi-layer perceptron to fulfil the undirected network representation task.

The results are summarized in Tables IV-V. From them, we indicate the following findings:

a) **M1 performs better in prediction accuracy among all involved models.** As shown in Table IV, on all data cases, M1 achieves better prediction accuracy than single-layered models. For instance, on C1.2, M1's RMSE is 0.1436, 4.58% lower than RMSE at 0.1505 by M2, 4.07% lower than RMSE at 0.1497 by M3, 5.71% lower than RMSE at 0.1523 by M4, and 16.6% lower than RMSE at 0.1722 by M5. On C3.1, M1's RMSE is 0.1263, 5.46% lower than M2's 0.1352, 5.46% lower than M3's 0.1319, 3.41% lower than M4's 0.1291, and 7.78% lower than M5's 0.1321. In comparison of M6, which is neural network-based, M1 achieves better prediction results in 8 of 10 cases. On the negative cases, the RMSE gap between M1 and M6 is 3.2% on C4.1 and 3.8% on C4.2. Experiment results indicate that proposed model has obvious advantage in prediction accuracy. The reasons are two-fold:

TABLE IV
PERFORMANCE OF INVOLVED MODELS ON DIFFERENT DATA CASES

| Data case | M1 | M2 | M3 | M4 | M5 | M6 |
|---|---|---|---|---|---|---|
| C1.1 | **0.1613±0.0079** | 0.1669±0.0080 | 0.1637±0.0075 | 0.1623±0.0077 | 0.1826±0.0062 | 0.1962±0.0045 |
| C1.2 | **0.1436±0.0057** | 0.1505±0.0067 | 0.1497±0.0052 | 0.1523±0.0059 | 0.1722±0.0042 | 0.1932±0.0058 |
| C2.1 | **0.1443±0.0034** | 0.1548±0.0098 | 0.1502±0.0076 | 0.1491±0.0087 | 0.1640±0.0054 | 0.1687±0.0049 |
| C2.2 | **0.1241±0.0028** | 0.1351±0.0066 | 0.1301±0.0051 | 0.1378±0.0064 | 0.1557±0.0070 | 0.1668±0.0025 |
| C3.1 | **0.1263±0.0077** | 0.1352±0.0089 | 0.1319±0.0057 | 0.1291±0.0070 | 0.1321±0.0047 | 0.1369±0.0014 |
| C3.2 | **0.1198±0.0033** | 0.1297±0.0047 | 0.1246±0.0031 | 0.1247±0.0046 | 0.1261±0.0034 | 0.1294±0.0026 |
| C4.1 | 0.3127±0.0040 | 0.4158±0.0064 | 0.3337±0.0042 | 0.3492±0.0074 | 0.4383±0.0039 | **0.3026±0.0021** |
| C4.2 | 0.2941±0.0039 | 0.4037±0.0036 | 0.3153±0.0047 | 0.3347±0.0048 | 0.4083±0.0076 | **0.2829±0.0039** |
| C5.1 | **0.0923±0.0021** | 0.1352±0.0043 | 0.1152±0.0050 | 0.1137±0.0072 | 0.1028±0.0017 | 0.0950±0.0013 |
| C5.2 | **0.0872±0.0048** | 0.1194±0.0033 | 0.0913±0.0033 | 0.0942±0.0047 | 0.0997±0.0013 | 0.0955±0.0042 |

i) M1 utilizes second order method as its optimizer, which is able to acquire inexact Newton steps for the train process. Note that the objective functions of SNLF models are non-convex, assimilate second order information into the train process can improve the model's representation ability by acquiring more accurate NLFs.

ii) Owing to the symmetric design and non-negative constraints incorporating strategy, M1 is able to maintain the key features of the input SHDI matrix during the training process since the second-order-based optimizers won't be disturbed by constraints. It makes great strides in maximizing utilization of undirected networks' known edges.

b) **M1 converge with affordable computational burden**. Although M1 utilize second order method, which is second-order-based, as their optimizer, their time consumption to converge is maintained at an acceptable level. For example, comparing to M3 and M4, which are SNLF models, M1 takes 14376 milliseconds to converge on C1.2, which is about 0.78 times and 0.57 times compared with M3's 18548 and M4's 25291, respectively.

Comparing to M2 and M5, which are matrix factorization-based and M6, which is neural network-based, the advantage of M1 in efficiency is obvious. On C1.2, M2, M5 and M6 takes 46052, 290184 and 1460147 milliseconds to converge, which is about 3.2, 20.2 and 101.6 times as much as M1's 14376 milliseconds. It is worth noting that even beaten by M6 in some cases, M1 outperforms it in achieving a better prediction accuracy and efficiency tradeoff.

TABLE V
AVERAGE TIME COSTS OF INVOLVED MODELS (MS)

| Data case | M1 | M2 | M3 | M4 | M5 | M6 |
|---|---|---|---|---|---|---|
| C1.1 | 11110 | 46052 | 1742 | 2505 | 290184 | 1460147 |
| C1.2 | 14376 | 277117 | 18548 | 25291 | 127891 | 1028798 |
| C2.1 | 9836 | 24281 | 1523 | 2793 | 438541 | 1628565 |
| C2.2 | 15382 | 65262 | 17558 | 22751 | 537826 | 5153501 |
| C3.1 | 4262 | 14823 | 628 | 1075 | 52819 | 730954 |
| C3.2 | 6468 | 93819 | 9691 | 11684 | 40563 | 2061678 |
| C4.1 | 103 | 259 | 16 | 25 | 1726 | 2747 |
| C4.2 | 225 | 297 | 147 | 108 | 9034 | 7703 |
| C5.1 | 572 | 2598 | 116 | 164 | 144485 | 58148 |
| C5.2 | 832 | 2734 | 582 | 630 | 40612 | 82762 |

V. CONCLUSIONS

Solving large-scale non-convex optimization problem is one of the core tasks for building SNLF models. Unfortunately, the performance of gradient-based solvers is limited since the SNLF model's objective function is high-dimensional and non-convex. Second-order approaches are expected to make a breakthrough. However, methods utilizing second order information directly are impractical due to their high computational cost. Moreover, the potential negative curvature problem and the requirement of maintaining NLFs' non-negativity make it difficult to acquire the NLFs with the improved second-order methods directly.

To address these issues, we propose $S^2NLF$, which adopts implicit symmetry and non-negative constraint modeling method by separating the output NLFs from the decision parameters with a mapping function that can always maintain the non-negativity of NLFs and integrate second-order information during the training process without utilizing the Hessian matrix directly. In comparison with state-of-the-art models, $S^2NLF$ performs well in prediction accuracy while remain the computational cost at an affordable level.